\documentclass[10pt]{article}

\usepackage[accepted]{tmlr}

\usepackage{amsmath,amsfonts,bm}

\def\eqref#1{equation~\ref{#1}}

\def\1{\bm{1}}

\DeclareMathAlphabet{\mathsfit}{\encodingdefault}{\sfdefault}{m}{sl}
\SetMathAlphabet{\mathsfit}{bold}{\encodingdefault}{\sfdefault}{bx}{n}

\DeclareMathOperator*{\argmax}{arg\,max}

\newcommand{\boldI}{{\boldsymbol{I}}}

\newcommand{\boldS}{{\boldsymbol{S}}}
\newcommand{\boldT}{{\boldsymbol{T}}}
\newcommand{\boldU}{{\boldsymbol{U}}}
\newcommand{\boldV}{{\boldsymbol{V}}}
\newcommand{\boldW}{{\boldsymbol{W}}}
\newcommand{\boldX}{{\boldsymbol{X}}}
\newcommand{\boldY}{{\boldsymbol{Y}}}

\newcommand{\boldh}{{\boldsymbol{h}}}

\newcommand{\boldx}{{\boldsymbol{x}}}
\newcommand{\boldy}{{\boldsymbol{y}}}

\usepackage{hyperref}
\usepackage{url}
\usepackage{graphicx, color}

\usepackage[noorphans]{quoting}
\usepackage[square,sort,comma,numbers]{natbib}

\usepackage{amsmath}
\usepackage{amssymb}
\usepackage{mathtools}
\usepackage{amsthm}
\usepackage{bbm}
\usepackage{booktabs}

\usepackage[most]{tcolorbox}

\usepackage[linesnumbered,ruled,vlined]{algorithm2e}
\SetKwInput{KwInput}{Input}
\SetKwInput{KwOutput}{Output}

\theoremstyle{plain}
\newtheorem{theorem}{Theorem}[section]
\newtheorem{proposition}[theorem]{Proposition}

\theoremstyle{definition}
\newtheorem{definition}[theorem]{Definition}

\theoremstyle{remark}

\newcommand*{\lbb}{\{\mskip-5mu\{}
\newcommand*{\rbb}{\}\mskip-5mu\}}

\usepackage{listings}
\title{Training-free Graph Neural Networks and the Power of \\ Labels as Features}

\author{\name Ryoma Sato \email rsato@nii.ac.jp \\
      \addr National Institute of Informatics}

\def\month{8}  
\def\year{2024} 
\def\openreview{\url{https://openreview.net/forum?id=7DzU88VrNU}} 

\begin{document}

\maketitle

\begin{abstract}
  We propose training-free graph neural networks (TFGNNs), which can be used without training and can also be improved with optional training, for transductive node classification. We first advocate labels as features (LaF), which is an admissible but not explored technique. We show that LaF provably enhances the expressive power of graph neural networks. We design TFGNNs based on this analysis. In the experiments, we confirm that TFGNNs outperform existing GNNs in the training-free setting and converge with much fewer training iterations than traditional GNNs.
\end{abstract}

\section{Introduction}

Graph Neural Networks (GNNs) \cite{gori2005new, scarselli2009graph} are popular machine learning models for processing graph data. GNNs show strong empirical performance in various machine learning and data mining tasks, including chemical modeling \cite{gilmer2017neural, jo2022score}, question answering \cite{tian2024graph,park2019estimating}, and recommender systems \cite{ying2018graph, wang2019knowledge, wans2019kgat, fan2019graph, hw2020light}.

One of the standard problem settings for GNNs is transductive node classification, where the goal is to predict the labels of the test nodes in a graph given the labels of other nodes. This setting has many applications, including document classification \cite{yang2016revisiting, kipf2017semi}, e-commerce \cite{schur2018pitfalls, zeng2020graph}, and social analysis \cite{hamilton2017inductive, zeng2020graph}. Many GNNs, including Graph Convolutional Networks (GCNs) \cite{kipf2017semi} and Graph Attention Networks (GATs) \cite{velickovic2018graph}, tackled transductive node classification and showed excellent performance.

One of the challenges of GNNs is the computational cost. There are many huge graphs in practice, such as social networks and Web graphs, which contain billions of nodes. It is sometimes prohibitive to even scan the entire graph, e.g., the whole World Wide Web. Many methods to speed up GNNs have been proposed. The basic approach is sampling nodes and/or edges to reduce the graph size \cite{hamilton2017inductive,chen2018fastgcn,zou2019layer,zeng2020graph}. In the extreme case, Sato et al. \cite{sato2022constant} proposed constant time graph neural networks by neighbor sampling. Although it drastically reduces the computational cost per iteration, it still requires many training iterations. PinSAGE \cite{ying2018graph} adopts parallel training with MapReduce as well as importance pooling to speed up the training. Although PinSAGE succeeded in training GNNs with Web-scale graphs, it requires massive computational resources. It is still challenging to instantly use GNNs with limited computational resources.

In this paper, we propose training-free graph neural networks (TFGNNs). 

To design TFGNNs, we first introduce the idea of labels as features (LaF). LaF uses the node labels as features, which is admissible in the framework of transductive node classification. GNNs with LaF can utilize the label information, such as the class distribution in the neighboring nodes, to compute the node embeddings, which contain much more information than the embeddings with only the node features. We show that LaF provably enhances the expressive power of GNNs.

TFGNNs can be used without training and deployed instantly as soon as the model is initialized. This property reduces the burden of hyperparameter tuning as no training process is involved in this mode. TFGNNs can also be improved with optional training. Users can use TFGNNs with the training-free mode or train TFGNNs for few iterations when the computational resources for training are limited. This property is useful for online learning, where training data come in a streaming manner, and the model should be updated instantly. Users can also fully train TFGNNs when the computational resources are abundant or the accuracy is required. TFGNNs enjoy the best of both worlds of nonparametric models and GNNs.

In the experiments, we confirm that TFGNNs outperform existing GNNs in the training-free setting and converge with much fewer training iterations than traditional GNNs. 

The contributions of this paper are as follows:

\begin{itemize}
  \item We advocate the use of LaF in transductive learning.
  \item We formally show that LaF strengthens the expressive power of GNNs.
  \item We proposed training-free graph neural networks (TFGNNs).
  \item We confirm that TFGNNs outperform existing GNNs in the training-free setting.
\end{itemize}

\begin{tcolorbox}[colframe=gray!20,colback=gray!20,sharp corners]
  \textbf{Reproducibility}: Our code is available at \url{https://github.com/joisino/laf}.
\end{tcolorbox}

\section{Background}

\subsection{Notations}

For every positive integer $n \in \mathbb{Z}_+$, $[n]$ denotes the set $\{1, 2, \ldots, n\}$. A graph is defined as a tuple of (i) the set $V$ of nodes, (ii) the set $E$ of edges, and (iii) the node features $\boldX = [\boldx_1, \boldx_2, \ldots, \boldx_n]^\top \in \mathbb{R}^{n \times d}$. Without loss of generality, we assume that the nodes are numbered with $1, 2, \ldots, n$. $\mathcal{Y}$ denotes the set of labels. $\boldy_v \in \mathbb{R}^{\mathcal{Y}}$ denotes the one-hot encoding of the label of node $v$. For every node $v \in V$, $\mathcal{N}(v)$ denotes the set of neighbors of node $v$. We adopt the numpy-like notation for indexing. For example, $\boldX_{:, 1}$ denotes the first column of $\boldX$, $\boldX_{:, -1}$ denotes the last column of $\boldX$, $\boldX_{:, -5:}$ denotes the last five columns of $\boldX$, and $\boldX_{:, :-5}$ denotes all the columns except for the last five columns of $\boldX$. 

\subsection{Transductive Node Classification}

\begin{tcolorbox}[colframe=gray!20,colback=gray!20,sharp corners]
  \textbf{Problem (Transductive Node Classification).} \\
  \textbf{Input}: A graph $G = (V, E, \boldX)$, labelled nodes $V_{\text{train}} \subset V$, and node labels $\boldY_{\text{train}} \in \mathcal{Y}^{V_{\text{train}}}$ of $V_{\text{train}}$. \\
  \textbf{Output}: Predicted labels $\boldY_{\text{test}} \in \mathcal{Y}^{V_{\text{test}}}$ of $V_{\text{test}} = V \setminus V_{\text{train}}$.
\end{tcolorbox}

There are two settings for the node classification problem: transductive and inductive. In transductive node classification, one graph and the labels of some of its nodes are given, and we predict the labels of the remaining nodes. This setting uses the same graph for training and testing. This is in contrast to the inductive setting, which uses different graphs for training and testing. For example, in spam account detection, annotating spam accounts on Facebook and using the trained model on Facebook is an example of the transductive setting, while using the trained model on X (Twitter) is an example of the inductive setting.

Transductive node classification is a popular setting in the GNN community; it has been employed in well-known studies, such as GCNs \cite{kipf2017semi} and GATs \cite{velickovic2018graph}, and has been adopted in popular benchmarks such as Cora, PubMed, and CiteSeer. There are also many practical applications of transductive node classification, such as document classification \cite{yang2016revisiting,kipf2017semi} and fraud detection\cite{wang2019semi,liu2019geniepath,liu2018heterogeneous}.

\subsection{Graph Neural Networks}

GNNs are a popular solution for transductive node classification. We follow the message-passing framework of GNNs \cite{gilmer2017neural}. A message passing GNN is defined as follows: \begin{align}
  \boldh^{(0)}_v &= \boldx_v & (\forall v \in V), \label{eq: gnn-def-init} \\
  \boldh^{(l)}_v &= f^{(l)}_{\text{agg}}(\boldh^{(l-1)}_v, \lbb \boldh^{(l-1)}_u \mid u \in \mathcal{N}(v) \rbb) & (\forall l \in [L], v \in V), \label{eq: gnn-def-agg} \\
  \hat{\boldy}_v &= f_{\text{pred}}(\boldh^{(L)}_v) & (\forall v \in V), \label{eq: gnn-def-pred}
\end{align} where $f^{(l)}_{\text{agg}}$ is the aggregation function and $f_{\text{pred}}$ is the prediction head, which are typically modeled by neural networks. 

\section{LaF is Admissible, but Not Explored Well} \label{sec: laf}

We ask the readers to recall the problem setting of transductive node classification. We are given node labels $\boldy_v$ of the training nodes. A typical approach for this problem is to feed node features $\boldx_v$ for a training node $v$ to the model, predict the labels of node $v$, compute the loss based on the ground truth label $\boldy_v$, and update the model parameters. However, how we use $\boldy_v$ is not limited. We can use $\boldy_v$ as features of node $v$ as well as for the loss function. This is the idea of LaF.

GNNs with LaF initialize the node embeddings in Eq. (\ref{eq: gnn-def-init}) as \begin{align}
  \boldh^{(0)}_v = [\boldx_v; \tilde{\boldy}_v] \in \mathbb{R}^{d + 1 + |\mathcal{Y}|}, \label{eq: gnn-laf-init}
\end{align} where $[\cdot; \cdot]$ denotes the concatenation of vectors, and \begin{align}
  \tilde{\boldy}_v = \begin{cases}
    [1; \boldy_v] & (v \in V_{\text{train}}), \\
    \bold0_{1 + |\mathcal{Y}|} & (v \in V_{\text{test}}),
  \end{cases}
\end{align} is the label vector of node $v$, and $\bold0_d$ is the zero vector of dimension $d$. LaF enables GNNs to utilize the label information, such as the class distribution in the neighboring nodes, to compute the node embeddings. Such embeddings are expected to be more informative than the embeddings without the label information. LaF is admissible in the sense that it uses only the information available in the transductive setting.

We emphasize that LaF has not been explored well in the literature on GNNs, regardless of its simplicity, with some notable exceptions \cite{wang2021bag, addanki2021large} (see Section \ref{sec: related} for detailed discussions). For example, GCNs \cite{kipf2017semi} and GATs \cite{velickovic2018graph} adopt the transductive setting, and they are allowed to use the label information as features. However, they initialize the node embeddings as $\boldh^{(0)}_v = \boldx_v$ without using the label information. One of the contributions of this paper is that we affirm that LaF is allowed in the transductive setting.

We should be careful when training GNNs with LaF. LaF may harm the generalization performance by inducing a shortcut of copying the label feature $\boldh_{v, d+1:}^{(0)}$ to the prediction. To prevent this, we should remove the label of the center nodes in the minibatch and treat them as test nodes. Specifically, let $B \subset V_{\text{train}}$ be the set of nodes in the minibatch and we set \begin{align}
  \tilde{\boldy}_v = \begin{cases}
    [1; \boldy_v] & (v \in V_{\text{train}} \setminus B), \\
    \bold0_{1 + |\mathcal{Y}|} & (v \in V_{\text{test}} \cup B),
  \end{cases}
\end{align} and predict the label $\hat{\boldy}_v$ for $v \in B$, and compute the loss based on $\hat{\boldy}_v$ and $\boldy_v$. This simulates the transductive setting where the label information of the test nodes is missing, and GNNs learn how to predict the labels of the test nodes based on the label information and node features of the surrounding nodes.

\section{LaF Strengthens the Expressive Power of GNNs} \label{sec: laf-power}

We show that LaF provably strengthens the expressive power of GNNs. Specifically, we show that GNNs with LaF can represent label propagation \cite{zhu2002learning}, an important model for transductive node classification, while GNNs without LaF cannot. This result is interesting in its own right, and it also motivates the design of TFGNNs.

Label propagation is a classic method for transductive node classification. It starts random walks from a test node and outputs the label distribution of the labeled nodes the random walks first hit. The following theorem shows that GNNs with LaF can represent label propagation.

\begin{theorem} \label{thm: laf}
  GNNs with LaF can approximate label propagation with any precision. Specifically, there exists a series of GNNs $\{f^{(l)}_{\text{agg}}\}_l$ and $f_{\text{pred}}$ such that for any positive $\varepsilon$, for any connected graph $G = (V, E, \boldX)$, for any labeled nodes $V_{\text{train}} \subset V$ and node labels $\boldY_{\text{train}} \in \mathcal{Y}^{V_{\text{train}}}$ and test node $v \in V \setminus V_{\text{train}}$, there exists $L \in \mathbb{Z}_+$ such that $l (\ge L)$-th GNN $(f^{(1)}_{\text{agg}}, \ldots, f^{(l)}_{\text{agg}}, f_{\text{pred}})$ with LaF outputs the approximation of label propagation with the error at most $\varepsilon$, i.e., \begin{align}
    \left\| \hat{\boldy}_v - \hat{\boldy}^{\text{LP}}_{v} \right\|_1 \le \varepsilon,
  \end{align} where $\hat{\boldy}^{\text{LP}}$ is the output of label propagation for test node $v$.
\end{theorem}

\begin{tcolorbox}[colframe=gray!10,colback=gray!10,sharp corners,breakable]
  \begin{proof}
    We prove the theorem by construction. Let \begin{align}
      p_{l, v, i} \stackrel{\text{def}}{=} \text{Pr}[\text{The random walk from node $v$ hits $V_{\text{train}}$ within $l$ steps and the first hit label is $i$}]. \label{eq: laf-p-def}
    \end{align} For the labeled nodes, this is a constant, i.e., \begin{align}
      p_{l, v, i} = 1[i = \boldy_v] \quad (\forall l \in \mathbb{Z}_{\ge 0}, v \in V_{\text{train}}, i \in \mathcal{Y}). \label{eq: laf-p-train}
    \end{align} For the other nodes, this can be recursively computed as follows: \begin{align}
      &p_{0, v, i} = 0 \quad (\forall v \in V \setminus V_{\text{train}}, i \in \mathcal{Y}), \label{eq: laf-rec-0} \\
      &p_{l, v, i} \notag \\
      \begin{split} &= \sum_{u \in \mathcal{N}(v)} \text{Pr}[\text{The first step is $v \to u$}] \cdot \text{Pr}[\text{The random walk from node $v$ hits $V_{\text{train}}$ within} \\ \\[-0.4in] & \hspace{2.1in} \text{ $l$ steps and the first hit label is $i$} \mid \text{The first step is $v \to u$}] \end{split} \\
      \begin{split} &= \sum_{u \in \mathcal{N}(v)} \frac{1}{\text{deg}(v)} \cdot \text{Pr}[\text{The random walk from node $u$ hits $V_{\text{train}}$ within $(l - 1)$ steps} \\ \\[-0.4in] & \hspace{1.3in} \text{and the first hit label is $i$}]\end{split} \\
      &= \frac{1}{\text{deg}(v)} \sum_{u \in \mathcal{N}(v)} p_{l - 1, u, i}. \label{eq: laf-rec}
    \end{align} These equations can be represented by GNNs with LaF. Specifically, the base case \begin{align}
    p_{0, v, i} = \begin{cases}
      1[i = \boldy_v] & (v \in V_{\text{train}}), \\
      0 & (v \in V \setminus V_{\text{train}}),
    \end{cases}
  \end{align} can be computed from $\tilde{\boldy}_v$ in $\boldh^{(0)}_v$. Let $f^{(l)}_{\text{agg}}$ always concat the first argument (i.e., $\boldh^{(l-1)}_v$ in Eq. (\ref{eq: gnn-def-agg})) to the output so that the GNN can keep the information of the input. $f^{(l)}_{\text{agg}}$ handles two cases by $\tilde{\boldy}_{v, 1} \in \{0, 1\}$, i.e., whether $v$ is in $V_{\text{train}}$ or not. If $v$ is in $V_{\text{train}}$, $f^{(l)}_{\text{agg}}$ just outputs $1[i = \boldy_v]$, which can be computed by $\tilde{\boldy}_v$ in $\boldh^{(l-1)}_v$. If $v$ is not in $V_{\text{train}}$, $f^{(l)}_{\text{agg}}$ aggregates $p_{l - 1, u, i}$ from $u$ in $\mathcal{N}(v)$ and takes the average, i.e., Eq. (\ref{eq: laf-rec}), which can be realized by message passing in the second argument of $f^{(l)}_{\text{agg}}$. The final output of the GNN is $p_{l, v, i}$. The output of label propagation can be decomposed as follows: \begin{align}
    &\hat{\boldy}^{\text{LP}}_{v, i} \notag \\
    &= \text{Pr}[\text{The first hit label is $i$}] \\
    \begin{split} &= \text{Pr}[\text{The random walk from node $v$ hits $V_{\text{train}}$ within $l$ steps and the first hit label is $i$}] \\ & \quad + \text{Pr}[\text{The random walk from node $v$ does not hit $V_{\text{train}}$ within $l$ steps and the first hit label is $i$}] \end{split} \\
    \begin{split} &= p_{l, v, i} \\ & \quad + \text{Pr}[\text{The random walk from node $v$ does not hit $V_{\text{train}}$ within $l$ steps and the first hit label is $i$}] \end{split}
  \end{align} As the second term converges to zero as $l$ increases, the GNNs approximate label propagation with any precision by increasing $l$.
\end{proof}
\end{tcolorbox}

We then show that GNNs without LaF cannot represent label propagation. 

\begin{proposition} \label{prop: no-laf}
  GNNs without LaF cannot approximate label propagation. Specifically, for any series of GNNs $\{f^{(l)}_{\text{agg}}\}_l$ and $f_{\text{pred}}$, there exists positive $\varepsilon$, a connected graph $G = (V, E, \boldX)$, labelled nodes $V_{\text{train}} \subset V$, node labels $\boldY_{\text{train}} \in \mathcal{Y}^{V_{\text{train}}}$ and test node $v \in V \setminus V_{\text{train}}$, such that for any $l$, GNN $(f^{(1)}_{\text{agg}}, \ldots, f^{(l)}_{\text{agg}}, f_{\text{pred}})$ without LaF has the error at least $\varepsilon$, i.e., \begin{align}
    \left\| \hat{\boldy}_v - \hat{\boldy}^{\text{LP}}_{v} \right\|_1 \ge \varepsilon,
  \end{align} where $\hat{\boldy}^{\text{LP}}$ is the output of label propagation for test node $v$.
\end{proposition}

\begin{tcolorbox}[colframe=gray!10,colback=gray!10,sharp corners]
  \begin{proof}
    We construct a counterexample. Let $G$ be a cycle of four nodes. The nodes are numbered as $1, 2, 3, 4$ in the clockwise direction. All the nodes have the same node features $\boldx$. Let $V_{\text{train}} = \{1, 2\}$ and $\boldY_{\text{train}} = [1, 0]^\top$. Label propagation classifies node $4$ as class $1$ and node $3$ as class $0$. However, GNNs without LaF always predict the same label for nodes $3$ and $4$ since they are isomorphic. Therefore, for any GNN without LaF, there is an irreducible error either for node $3$ or $4$. 
  \end{proof}
\end{tcolorbox}

Theorem \ref{thm: laf} and Proposition \ref{prop: no-laf} show that LaF provably enhances the expressive power of GNNs. These results indicate that GNNs with LaF are more powerful than traditional message passing GNNs such as GCNs, GATs, and GINs without LaF. Note that GINs have been considered to be the most expressive message passing GNNs, but GINs cannot represent label propagation without LaF while message passing GNNs with LaF can. This does not lead to a contradiction since the original GINs do not take the label information as input. Put differently, the input domains of the functions differ. These results indicate that it is important to consider what to input to GNNs as well as the architecture of GNNs.

\section{Training-free Graph Neural Networks} \label{sec: tfgnn}

We propose training-free graph neural networks (TFGNNs) based on the analysis in the previous section. TFGNNs can be used without training and can also be improved with optional training.

First, we define training-free models.

\begin{definition}[Training-free Model] \label{def: training-free}
  We say a parametric model is training-free if it can be used without optimizing the parameters.
\end{definition}

It should be noted that nonparametric models are training-free by definition. The real worth of TFGNNs is that it is training-free while it can be improved with optional training. Users can enjoy the best of both worlds of parametric and nonparametric models by choosing the trade-off based on the computational resources for training and the accuracy required.

\begin{figure}[tb]
  \centering
  \includegraphics[width=0.6\hsize]{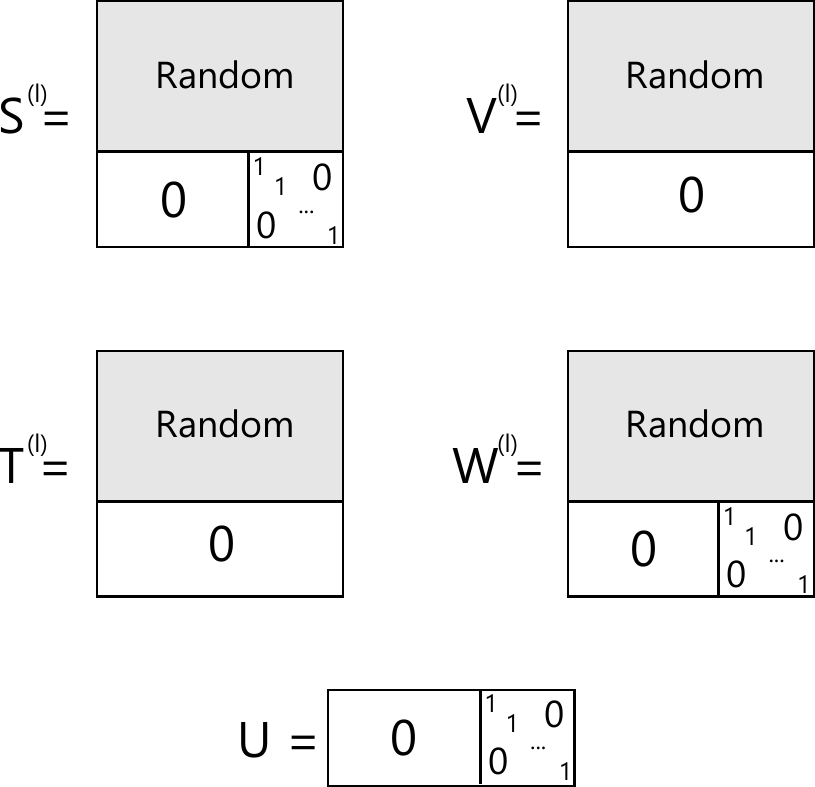}
  \caption{Initialization of TFGNNs. The parameters of the last $(1 + |\mathcal{Y}|)$ rows or $|\mathcal{Y}|$ rows are initialized by $0$ or $1$ in a special pattern}
  \label{fig: init}
\end{figure}

The core idea of TFGNNs is to embed label propagation in GNNs by Theorem \ref{thm: laf}. TFGNNs are defined as follows: \begin{align}
  \boldh^{(0)}_v &= [\boldx_v; \tilde{\boldy}_v] & (\forall v \in V), \label{eq: tfgnn-def-init} \\
  \boldh^{(l)}_v &= \begin{cases}
    \text{ReLU}\left(\boldS^{(l)} \boldh^{(l-1)}_v + \frac{1}{|\mathcal{N}(v)|} \sum_{u \in \mathcal{N}(v)} \boldV^{(l)} \boldh^{(l-1)}_u\right) & (v \in V_{\text{train}}, l \in [L]), \\
    \text{ReLU}\left(\boldT^{(l)} \boldh^{(l-1)}_v + \frac{1}{|\mathcal{N}(v)|} \sum_{u \in \mathcal{N}(v)} \boldW^{(l)} \boldh^{(l-1)}_u\right) & (v \in V_{\text{test}}, l \in [L]),
  \end{cases} \\
  \hat{\boldy}_v &= \text{softmax}(\boldU \boldh^{(L)}_v) & (\forall v \in V).
\end{align} The architecture of TFGNNs is standard, i.e., TFGNNs transform the center nodes and carry out mean aggregation from the neighboring nodes. The key to TFGNNs lies in initialization. The parameters are initialized as follows: \begin{align}
  \boldS^{(l)}_{-(1 + |\mathcal{Y}|):, :-(1 + |\mathcal{Y}|)} &= 0 & (\forall l \in [L]), \\
  \boldS^{(l)}_{-(1 + |\mathcal{Y}|):, -(1 + |\mathcal{Y}|):} &= \boldI_{1 + |\mathcal{Y}|} & (\forall l \in [L]), \\
  \boldV^{(l)}_{-(1 + |\mathcal{Y}|):} &= 0 & (\forall l \in [L]), \\
  \boldT^{(l)}_{-(1 + |\mathcal{Y}|):} &= 0 & (\forall l \in [L]), \\
  \boldW^{(l)}_{-(1 + |\mathcal{Y}|):, :-(1 + |\mathcal{Y}|)} &= 0 & (\forall l \in [L]), \\
  \boldW^{(l)}_{-(1 + |\mathcal{Y}|):, -(1 + |\mathcal{Y}|):} &= \boldI_{1 + |\mathcal{Y}|} & (\forall l \in [L]), \\
  \boldU_{:, :-|\mathcal{Y}|} &= 0, \\
  \boldU_{:, -|\mathcal{Y}|:} &= \boldI_{|\mathcal{Y}|}, \label{eq: tfgnn-init-last}
\end{align} i.e., the parameters of the last $(1 + |\mathcal{Y}|)$ rows or $|\mathcal{Y}|$ rows are initialized by $0$ or $1$ in a special pattern (Figure \ref{fig: init}). Other parameters are initialized randomly, e.g., by Xavier initialization \cite{glorot2010understanding}. The following proposition shows that the initialized TFGNNs approximate label propagation.

\begin{proposition} \label{prop: tfgnn}
  The initialized TFGNNs approximate label propagation. Specifically, \begin{align}
    \boldh^{(L)}_{v, -(|\mathcal{Y}| - i + 1)} = p_{L, v, i}
  \end{align} holds, where $p_{L, v, i}$ is defined in Eq. (\ref{eq: laf-p-def}), and \begin{align}
    \argmax_i \hat{\boldy}_{vi} = \argmax_i p_{L, v, i}
  \end{align} holds, and $p_{L, v, i} \to \hat{\boldy}^{\text{LP}}_{v, i}$ as $L \to \infty$.
\end{proposition}

\begin{tcolorbox}[colframe=gray!10,colback=gray!10,sharp corners]
  \begin{proof}
    By the definitions of TFGNNs, \begin{align}
      \boldh^{(0)}_{v, -|\mathcal{Y}|:} &= \begin{cases}
        \boldy_v & (v \in V_{\text{train}}), \\
        \bold0_{|\mathcal{Y}|} & (v \in V_{\text{test}}),
      \end{cases} \\
      \boldh^{(l)}_{v, -|\mathcal{Y}|:} &= \begin{cases}
        \boldh^{(l-1)}_{v, -|\mathcal{Y}|:} & (v \in V_{\text{train}}, l \in [L]), \\
        \frac{1}{|\mathcal{N}(v)|} \sum_{u \in \mathcal{N}(v)} \boldh^{(l-1)}_{u, -|\mathcal{Y}|:} & (v \in V_{\text{test}}, l \in [L]).
      \end{cases}
    \end{align}
    This recursion is the same as Eqs. (\ref{eq: laf-p-train}) -- (\ref{eq: laf-rec}). Therefore, \begin{align}
      \boldh^{(L)}_{v, -(|\mathcal{Y}| - i + 1)} = p_{L, v, i}
    \end{align}
    holds. As $\boldU$ picks the last $|\mathcal{Y}|$ dimensions, and softmax is monotone, \begin{align}
      \argmax_i \hat{\boldy}_{vi} = \argmax_i p_{L, v, i}
    \end{align} holds. $p_{L, v, i} \to \hat{\boldy}^{\text{LP}}_{v, i}$ as $L \to \infty$ is shown in the proof of Theorem \ref{thm: laf}.
  \end{proof}
\end{tcolorbox}

Therefore, the initialized TFGNNs can be used for transductive node classification as are without training. The approximation algorithm of label propagation is seamlessly embedded in the model parameters, and TFGNNs can also be trained as usual GNNs.

\section{Experiments} \label{sec: exp}

\subsection{Experimental Setup}

We use the Planetoid datasets (Cora, CiteSeer, PubMed) \cite{yang2016revisiting}, Coauthor datasets, and Amazon datasets \cite{schur2018pitfalls} in the experiments. We use 20 nodes per class for training, 500 nodes for validation, and the rest for testing in the Planetoid datasets following Kipf et al. \cite{kipf2017semi}, and use 20 nodes per class for training, $30$ nodes per class for validation, and the rest for testing in the Coauthor and Amazon datasets following Shchur et al. \cite{schur2018pitfalls}. We use GCNs \cite{kipf2017semi} and GATs \cite{velickovic2018graph} for the baselines. We use three layered models with the hidden dimension $32$ unless otherwise specified. We train all the models with AdamW \cite{loshchilov2019decoupled} with learning rate $0.0001$ and weight decay $0.01$.

\subsection{TFGNNs Outperform Existing GNNs in Training-free Setting}

\begin{table}[t]
  \centering
  \caption{Node classification accuracy in the training-free setting. The best results are shown in \textbf{bold}. \\ CS: Coauthor CS, Physics: Coauthor Physics, Computers: Amazon Computers, Photo: Amazon Photo. TFGNNs outperform GCNs and GATs in all the datasets. These results indicate that TFGNNs are training-free. Note that we use three-layered TFGNNs to make the comparison fair although deeper TFGNNs perform better in the training-free setting as we confirm in Section \ref{sec: depth}.}
  \label{table: training-free}
  \scalebox{0.9}{%
  \begin{tabular}{lccccccc}
  \toprule
   & Cora & CiteSeer & PubMed & CS & Physics & Computers & Photo \\
  \midrule
  GCNs & 0.163 & 0.167 & 0.180 & 0.079 & 0.101 & 0.023 & 0.119 \\
  GCNs + LaF & 0.119 & 0.159 & 0.407 & 0.080 & 0.146 & 0.061 & 0.142 \\
  GATs & 0.177 & 0.229 & 0.180 & 0.040 & 0.163 & 0.058 & 0.122 \\
  GATs + LaF & 0.319 & 0.077 & 0.180 & 0.076 & 0.079 & 0.025 & 0.044 \\
  TFGNNs + random initialization & 0.149 & 0.177 & 0.180 & 0.023 & 0.166 & 0.158 & 0.090 \\
  TFGNNs (proposed) & \textbf{0.600} & \textbf{0.362} & \textbf{0.413} & \textbf{0.601} & \textbf{0.717} & \textbf{0.730} & \textbf{0.637} \\
  \bottomrule
  \end{tabular}
  }
\end{table}

We compare the performance of TFGNNs with GCNs and GATs in the training-free setting by assessing the accuracy of the models when the parameters are initialized. The results are shown in Table \ref{table: training-free}. TFGNNs outperform GCNs and GATs in all the datasets. Specifically, both GCNs and GATs are almost random in the training-free setting, while TFGNNs achieve non-trivial accuracy. These results validate that TFGNNs meet the definition \ref{def: training-free} of training-free models. We can also observe that GCNs, GATs, and TFGNNs do not benefit from LaF in the training-free settings if randomly initialized. These results indicate that both LaF and the initialization of TFGNNs are important for training-free performance.

\subsection{Deep TFGNNs Perform Better in Training-free Setting} \label{sec: depth}

\begin{figure}[tb]
  \centering
  \includegraphics[width=\hsize]{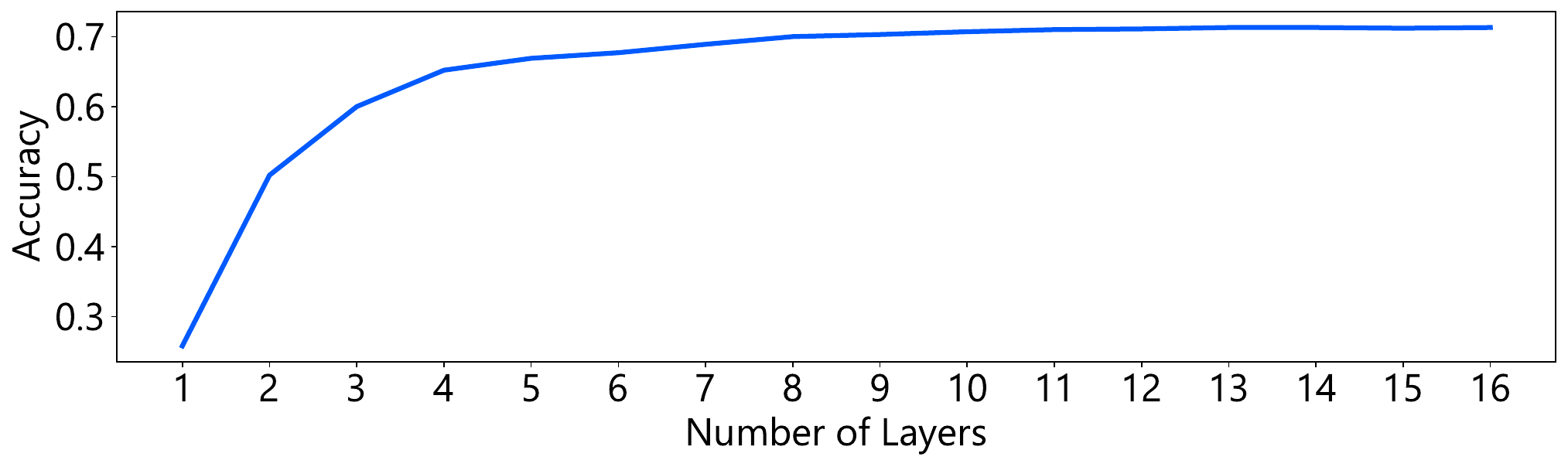}
  \caption{Deep TFGNNs perform better in the training-free setting. The x-axis is the number of layers, and the y-axis is the accuracy of the models for the Cora dataset in the training-free setting. These results show that deeper TFGNNs perform better in the training-free setting.}
  \label{fig: depth}
\end{figure}

We confirm that deeper TFGNNs perform better in the training-free setting. We have used three-layered TFGNNs so far to make the comparison fair with existing GNNs. Proposition \ref{prop: tfgnn} shows that the initialized TFGNNs converge to label propagation as the depth goes to infinity, and we expect that deeper TFGNNs perform better in the training-free setting. Figure \ref{fig: depth} shows the accuracy of TFGNNs with different depths for the Cora dataset. We can observe that deeper TFGNNs perform better in the training-free setting until the depth reaches around $10$, where the performance saturates. It is noteworthy that GNNs have been known to suffer from the oversmoothing problem \cite{li2018deeper, oono2020graph}, and the performance of GNNs degrades as the depth increases. It is interesting that TFGNNs do not suffer from the oversmoothing problem in the training-free setting. It should be noted that it does not necessarily mean that deeper models perform better in the optional training mode because the optional training may break the structure introduced by the initialization of TFGNNs and may lead to oversmoothing and/or overfitting. We leave it as a future work to overcome these problems by adopting countermeasures such as initial residual and identity mapping \cite{simple2020chen}, MADReg \cite{chen2020measuring}, and DropEdge \cite{rong2020dropedge}.

\subsection{TFGNNs Converge Fast}

\begin{figure}[tb]
  \centering
  \includegraphics[width=\hsize]{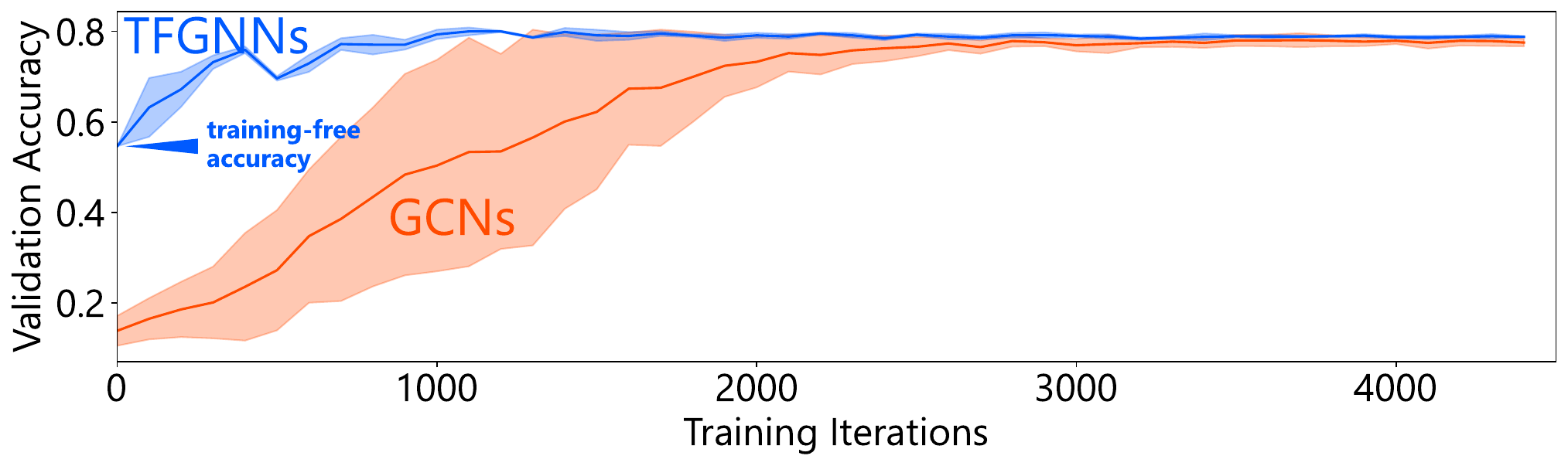}
  \caption{TFGNNs converge fast. The x-axis is the number of training iterations, and the y-axis is the validation accuracy of the models for the Cora dataset. These results show that TFGNNs in the optional training mode converge much faster than GCNs.}
  \label{fig: training-curve}
\end{figure}

\begin{figure}[tb]
  \centering
  \includegraphics[width=\hsize]{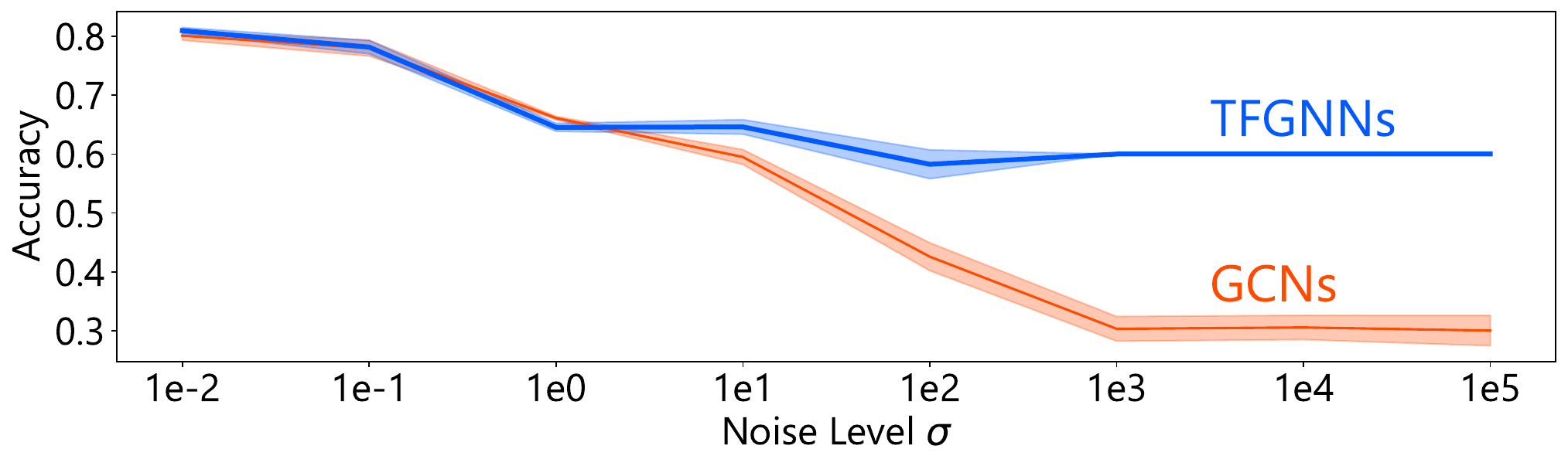}
  \caption{TFGNNs are robust to feature noise. The x-axis is the standard deviation of the Gaussian noise added to the node features, and the y-axis is the accuracy of the models for the Cora dataset. Both models are trained. These results show that TFGNNs are more robust to feature noise than GCNs.}
  \label{fig: noise}
\end{figure}

In the following, we investigate the optional training mode of TFGNNs. We train the models with three random seeds and report the average accuracy and standard deviation. We use baseline GCNs without LaF (i.e., the original GCNs) as the baseline.   

First, we confirm that TFGNNs in the optional training mode converge faster than GCNs. We show the training curves of TFGNNs and GCNs for the Cora dataset in Figure \ref{fig: training-curve}. TFGNNs converge much faster than GCNs. We hypothesize that TFGNNs converge faster because the initialized TFGNNs are in a good starting point, while GCNs start from a completely random point and require many iterations to reach a good point. We can also observe that fully trained TFGNNs perform on par with GCNs. These results indicate that TFGNNs enjoy the best of both worlds: TFGNNs perform well without training and can be trained faster with optional training.

\subsection{TFGNNs are Robust to Feature Noise}

As TFGNNs use both node features and label information while traditional GNNs rely only on node features, we expect that TFGNNs are more robust to feature noise than traditional GNNs. We confirm this in this section. We add i.i.d. Gaussian noise with standard deviation $\sigma$ to the node features and evaluate the accuracy of the models. We train TFGNNs and GCNs with the Cora dataset. The results are shown in Figure \ref{fig: noise}. TFGNNs are more robust to feature noise especially in high noise regimes where the performance of GCNs degrades significantly. These results indicate that TFGNNs are more robust to i.i.d. Gaussian noise to the node features than traditional GNNs.

\section{Related Work} \label{sec: related}

\subsection{Labels as Features and Training-free GNNs}

The most relevant work is \citet{wang2021bag}, who proposed to use node labels in GNNs. This technique was also used in \citet{addanki2021large} and analyzied in \citet{wang2022why}. The underlying idea is common with LaF, i.e., use of label information as input to transductive GNNs. A similar result as Theorem \ref{thm: laf} was also shown in \cite{wang2022why}. However, the focus is different, and there are different points between this work and theirs. We propose the training-free + optional training framework for the first time. The notable characteristics of GNNs are (i) TFGNNs receive both original features and LaF, (ii) TFGNNs can be deployed without training, and (iii) TFGNNs can be improved with optional training. Besides, we provide detailed analysis and experiments including the speed of convergence and noise robustness. Our results provide complementary insights to the existing works.

Another related topic is graph echo state networks \cite{gallicchio2010graph, gallicchio2020fast, micheli2023addressing}, which lead to lightweight models for graph data. The key idea is to use randomly initialized fixed weights for aggregation. The main difference is that graph echo state networks still require to train the output layer, while TFGNNs can be used without training. These methods are orthogonal, and it is an interesting direction to combine them to further improve the performance.

\subsection{Speeding up GNNs}

Various methods have been proposed to speed up GNNs to handle large graph data. GraphSAGE \cite{hamilton2017inductive} is one of the earliest methods to speed up GNNs. GraphSAGE employs neighbor sampling to reduce the computational cost of training and inference. It samples a fixed number of neighbors for each node and aggregates the features of the sampled neighbors. An alternative sampling method is layer-wise sampling introduced in FastGCN \cite{chen2018fastgcn}. Huang et al. \cite{huang2018adaptive} further improved FastGCN by using an adaptive node sampling technique to reduce the variance of estimators. LADIES \cite{zou2019layer} combined neighbor sampling and layer-wise sampling to take the best of both worlds. Another approach is to use smaller training graphs. ClusterGCN \cite{chiang2019cluster} uses a cluster of nodes as a mini-batch. GraphSAINT \cite{zeng2020graph} samples subgraphs by random walks for each mini-batch.

It should also be noted that general techniques to speed up neural networks, such as mixed-precision training \cite{micikevicius2018mixed}, quantization \cite{wang2018training,sun2020ultra,jacob2018quantization, wu2020integer, krishnamoorthi2018quantizing}, and pruning \cite{han2015learning,blalock2020what} can be applied to GNNs.

These methods mitigate the training cost of GNNs, but they still require many training iterations. In this paper, we propose training-free GNNs, which can be deployed instantly as soon as the model is initialized. Besides, our method can be improved with optional training. In the optional training mode, the speed up techniques mentioned above can be combined with our method to reduce the training time further.

\subsection{Expressive Power of GNNs}

Expressive power (or representation power) means what kind of functional classes a model family can realize. The expressive power of GNNs is an important field of research in its own right. If GNNs cannot represent the true function, we cannot expect GNNs to work well however we train them. Therefore, it is important to elucidate the expressive power of GNNs. Originally, Morris et al. \cite{morris2019weisfeiler} and Xu et al. \cite{xu2019how} showed that message-passing GNNs are at most as powerful as the 1-WL test, and they proposed $k$-GNNs and GINs, which are as powerful as the $k$-(set)WL and 1-WL tests, respectively. GINs are the most powerful message-passing GNNs. Sato \cite{sato2019approximation, sato2021random} and Loukas \cite{loukas2020what} showed that message-passing GNNs are as powerful as a computational model of distributed local algorithms, and they proposed GNNs that are as powerful as port-numbering and randomized local algorithms. Loukas \cite{loukas2020what} showed that GNNs are Turing-complete under certain conditions (i.e., with unique node ids and infinitely increasing depths). Some other works showed that GNNs can solve or cannot solve some specific problems, e.g., GNNs can recover the underlying geometry \cite{sato2023graph}, GNNs cannot recognize bridges and articulation points \cite{zhang2023rethinking}. There are various efforts to improve the expressive power of GNNs by non-message-passing architectures \cite{maron2019invariant, maron2019provably, murphy2019relational}. We refer the readers to survey papers \cite{sato2020survey, jegelka2022theory} for more details on the expressive power of GNNs.

We contributed to the field of the expressive power of GNNs by showing that GNNs with LaF are more powerful than GNNs without LaF. Specifically, we showed that GNNs with LaF can represent an important model, label propagation, while GNNs without LaF cannot. It should be emphasized that GINs, the most powerful message-passing GNNs, and Turing-complete GNNs cannot represent label propagation without LaF because they do not have access to the label information label propagation uses, and also noted that GINs traditionally do not use LaF. This result indicates that it is important to consider what to input to the GNNs as well as the architecture of the GNNs for the expressive power of GNNs. This result provides a new insight into the field of the expressive power of GNNs.

\section{Limitations}

Our work has several limitations. First, LaF and TFGNNs cannot be applied to inductive settings while most GNNs can. We do not regard this as a negative point. Popular GNNs such as GCNs and GATs are applicable to both transductive and inductive settings and are often used for transductive settings. However, this also means that they do not take advantage of transductive-specific structures (those that are not present in inductive settings). We believe that it is important to exploit inductive-specific techniques for inductive settings and transductive-specific techniques (such as LaF) for transductive settings in order to pursue maximum performance.

Second, TFGNNs cannot be applied to heterophilious graphs, or its performance degrades as TFGNNs are based on label propagation. The same argument mentioned above applies. Relying on homophilious graphs is not a negative point in pursuing maximum performance. It should be noted that LaF may also be exploited in heterophilious settings as well. Developing training-free GNNs for heterophilious graphs based on LaF is an interesting future work.

Third, we did not aim to achieve the state-of-the-art performance. Exploring the combination of LaF with fancy techniques to achieve state-of-the-art performance is left as future work.

Finally, we did not explore applications of LaF other than TFGNNs. LaF can help other GNNs in non-training-free settings as well. Exploring the application of LaF to other GNNs is left as future work.

\section{Conclusion}

In this paper, we made the following contributions.

\begin{itemize}
  \item We advocated the use of LaF in transductive learning (Section \ref{sec: laf}). \begin{itemize}
    \item We confirmed that LaF is admissible in transductive learning, but LaF has not been explored in the field of GNNs such as GCNs and GATs.
  \end{itemize}
  \item We formally showed that LaF strengthens the expressive power of GNNs (Section \ref{sec: laf-power}). \begin{itemize}
    \item We showed that GNNs with LaF can represent label propagation (Theorem \ref{thm: laf}) while GNNs without LaF cannot (Proposition \ref{prop: no-laf}).
  \end{itemize}
  \item We proposed training-free graph neural networks, TFGNNs (Section \ref{sec: tfgnn}). \begin{itemize}
  \item We showed that TFGNNs defined by Eqs. (\ref{eq: tfgnn-def-init}) -- (\ref{eq: tfgnn-init-last}) meet the requirements of training-free models (Definition \ref{def: training-free}) by showing that the initialized TFGNNs approximate label propagation in Proposition \ref{prop: tfgnn}.
  \end{itemize}
  \item We confirmed that TFGNNs outperform existing GNNs in the training-free setting. (Section \ref{sec: exp}) \begin{itemize}
    \item We showed that TFGNNs outperform GCNs and GATs in all of the seven datasets in the training-free setting (Table \ref{table: training-free}).
    \item TFGNNs achieve non-trivial accuracy without training and can be deployed instantly as soon as the model is initialized. These results corroborate that TFGNNs are training-free (Definition \ref{def: training-free}).
  \end{itemize}
\end{itemize}

We also note that our idea can be applied to other machine learning models than graph neural networks. We hope that this papers opens the door to a new research direction of training-free neural networks.

\bibliography{main}

\begin{thebibliography}{59}
\providecommand{\natexlab}[1]{#1}
\providecommand{\url}[1]{\texttt{#1}}
\expandafter\ifx\csname urlstyle\endcsname\relax
  \providecommand{\doi}[1]{doi: #1}\else
  \providecommand{\doi}{doi: \begingroup \urlstyle{rm}\Url}\fi

\bibitem[Addanki et~al.(2021)Addanki, Battaglia, Budden, Deac, Godwin, Keck,
  Li, Sanchez{-}Gonzalez, Stott, Thakoor, and Velickovic]{addanki2021large}
R.~Addanki, P.~W. Battaglia, D.~Budden, A.~Deac, J.~Godwin, T.~Keck, W.~L.~S.
  Li, A.~Sanchez{-}Gonzalez, J.~Stott, S.~Thakoor, and P.~Velickovic.
\newblock Large-scale graph representation learning with very deep gnns and
  self-supervision.
\newblock \emph{arXiv}, 2021.
\newblock URL \url{https://arxiv.org/abs/2107.09422}.

\bibitem[Blalock et~al.(2020)Blalock, Ortiz, Frankle, and
  Guttag]{blalock2020what}
D.~W. Blalock, J.~J.~G. Ortiz, J.~Frankle, and J.~V. Guttag.
\newblock What is the state of neural network pruning?
\newblock In \emph{Proceedings of Machine Learning and Systems 2020, MLSys},
  2020.

\bibitem[Chen et~al.(2020{\natexlab{a}})Chen, Lin, Li, Li, Zhou, and
  Sun]{chen2020measuring}
D.~Chen, Y.~Lin, W.~Li, P.~Li, J.~Zhou, and X.~Sun.
\newblock Measuring and relieving the over-smoothing problem for graph neural
  networks from the topological view.
\newblock In \emph{Proceedings of the 34th {AAAI} Conference on Artificial
  Intelligence, {AAAI}}, pages 3438--3445, 2020{\natexlab{a}}.

\bibitem[Chen et~al.(2018)Chen, Ma, and Xiao]{chen2018fastgcn}
J.~Chen, T.~Ma, and C.~Xiao.
\newblock {FastGCN}: Fast learning with graph convolutional networks via
  importance sampling.
\newblock In \emph{Proceedings of the 6th International Conference on Learning
  Representations, {ICLR}}, 2018.

\bibitem[Chen et~al.(2020{\natexlab{b}})Chen, Wei, Huang, Ding, and
  Li]{simple2020chen}
M.~Chen, Z.~Wei, Z.~Huang, B.~Ding, and Y.~Li.
\newblock Simple and deep graph convolutional networks.
\newblock In \emph{Proceedings of the 37th International Conference on Machine
  Learning, {ICML}}, pages 1725--1735, 2020{\natexlab{b}}.

\bibitem[Chiang et~al.(2019)Chiang, Liu, Si, Li, Bengio, and
  Hsieh]{chiang2019cluster}
W.~Chiang, X.~Liu, S.~Si, Y.~Li, S.~Bengio, and C.~Hsieh.
\newblock Cluster-gcn: An efficient algorithm for training deep and large graph
  convolutional networks.
\newblock In \emph{Proceedings of the 25th {ACM} {SIGKDD} International
  Conference on Knowledge Discovery {\&} Data Mining, {KDD}}, pages 257--266,
  2019.

\bibitem[Fan et~al.(2019)Fan, Ma, Li, He, Zhao, Tang, and Yin]{fan2019graph}
W.~Fan, Y.~Ma, Q.~Li, Y.~He, Y.~E. Zhao, J.~Tang, and D.~Yin.
\newblock Graph neural networks for social recommendation.
\newblock In \emph{The Web Conference 2019, {WWW}}, pages 417--426, 2019.

\bibitem[Gallicchio and Micheli(2010)]{gallicchio2010graph}
C.~Gallicchio and A.~Micheli.
\newblock Graph echo state networks.
\newblock In \emph{International Joint Conference on Neural Networks, {IJCNN}},
  pages 1--8, 2010.

\bibitem[Gallicchio and Micheli(2020)]{gallicchio2020fast}
C.~Gallicchio and A.~Micheli.
\newblock Fast and deep graph neural networks.
\newblock In \emph{Proceedings of the 34th {AAAI} Conference on Artificial
  Intelligence, {AAAI}}, pages 3898--3905, 2020.

\bibitem[Gilmer et~al.(2017)Gilmer, Schoenholz, Riley, Vinyals, and
  Dahl]{gilmer2017neural}
J.~Gilmer, S.~S. Schoenholz, P.~F. Riley, O.~Vinyals, and G.~E. Dahl.
\newblock Neural message passing for quantum chemistry.
\newblock In \emph{Proceedings of the 34th International Conference on Machine
  Learning, {ICML}}, pages 1263--1272, 2017.

\bibitem[Glorot and Bengio(2010)]{glorot2010understanding}
X.~Glorot and Y.~Bengio.
\newblock Understanding the difficulty of training deep feedforward neural
  networks.
\newblock In \emph{Proceedings of the Thirteenth International Conference on
  Artificial Intelligence and Statistics, {AISTATS}}, pages 249--256, 2010.

\bibitem[Gori et~al.(2005)Gori, Monfardini, and Scarselli]{gori2005new}
M.~Gori, G.~Monfardini, and F.~Scarselli.
\newblock A new model for learning in graph domains.
\newblock In \emph{Proceedings of the International Joint Conference on Neural
  Networks, {IJCNN}}, volume~2, pages 729--734, 2005.

\bibitem[Hamilton et~al.(2017)Hamilton, Ying, and
  Leskovec]{hamilton2017inductive}
W.~L. Hamilton, Z.~Ying, and J.~Leskovec.
\newblock Inductive representation learning on large graphs.
\newblock In \emph{Advances in Neural Information Processing Systems 30: Annual
  Conference on Neural Information Processing Systems 2017, {NeurIPS}}, pages
  1024--1034, 2017.

\bibitem[Han et~al.(2015)Han, Pool, Tran, and Dally]{han2015learning}
S.~Han, J.~Pool, J.~Tran, and W.~J. Dally.
\newblock Learning both weights and connections for efficient neural network.
\newblock In \emph{Advances in Neural Information Processing Systems 28: Annual
  Conference on Neural Information Processing Systems 2015, {NeurIPS}}, pages
  1135--1143, 2015.

\bibitem[He et~al.(2020)He, Deng, Wang, Li, Zhang, and Wang]{hw2020light}
X.~He, K.~Deng, X.~Wang, Y.~Li, Y.~Zhang, and M.~Wang.
\newblock Lightgcn: Simplifying and powering graph convolution network for
  recommendation.
\newblock In \emph{Proceedings of the 43rd International {ACM} {SIGIR}
  conference on research and development in Information Retrieval, {SIGIR}},
  pages 639--648, 2020.

\bibitem[Huang et~al.(2018)Huang, Zhang, Rong, and Huang]{huang2018adaptive}
W.~Huang, T.~Zhang, Y.~Rong, and J.~Huang.
\newblock Adaptive sampling towards fast graph representation learning.
\newblock In \emph{Advances in Neural Information Processing Systems 31,
  NeurIPS}, 2018.

\bibitem[Jacob et~al.(2018)Jacob, Kligys, Chen, Zhu, Tang, Howard, Adam, and
  Kalenichenko]{jacob2018quantization}
B.~Jacob, S.~Kligys, B.~Chen, M.~Zhu, M.~Tang, A.~G. Howard, H.~Adam, and
  D.~Kalenichenko.
\newblock Quantization and training of neural networks for efficient
  integer-arithmetic-only inference.
\newblock In \emph{2018 {IEEE} Conference on Computer Vision and Pattern
  Recognition, {CVPR}}, pages 2704--2713, 2018.

\bibitem[Jegelka(2022)]{jegelka2022theory}
S.~Jegelka.
\newblock Theory of graph neural networks: Representation and learning.
\newblock \emph{arXiv}, abs/2204.07697, 2022.

\bibitem[Jo et~al.(2022)Jo, Lee, and Hwang]{jo2022score}
J.~Jo, S.~Lee, and S.~J. Hwang.
\newblock Score-based generative modeling of graphs via the system of
  stochastic differential equations.
\newblock In \emph{International Conference on Machine Learning, {ICML}}, pages
  10362--10383, 2022.

\bibitem[Kipf and Welling(2017)]{kipf2017semi}
T.~N. Kipf and M.~Welling.
\newblock Semi-supervised classification with graph convolutional networks.
\newblock In \emph{Proceedings of the 5th International Conference on Learning
  Representations, {ICLR}}, 2017.

\bibitem[Krishnamoorthi(2018)]{krishnamoorthi2018quantizing}
R.~Krishnamoorthi.
\newblock Quantizing deep convolutional networks for efficient inference: {A}
  whitepaper.
\newblock \emph{arXiv}, abs/1806.08342, 2018.
\newblock URL \url{http://arxiv.org/abs/1806.08342}.

\bibitem[Li et~al.(2018)Li, Han, and Wu]{li2018deeper}
Q.~Li, Z.~Han, and X.~Wu.
\newblock Deeper insights into graph convolutional networks for semi-supervised
  learning.
\newblock In \emph{Proceedings of the 32nd {AAAI} Conference on Artificial
  Intelligence, {AAAI}}, pages 3538--3545, 2018.

\bibitem[Liu et~al.(2018)Liu, Chen, Yang, Zhou, Li, and
  Song]{liu2018heterogeneous}
Z.~Liu, C.~Chen, X.~Yang, J.~Zhou, X.~Li, and L.~Song.
\newblock Heterogeneous graph neural networks for malicious account detection.
\newblock In \emph{Proceedings of the 27th {ACM} International Conference on
  Information and Knowledge Management, {CIKM}}, pages 2077--2085, 2018.

\bibitem[Liu et~al.(2019)Liu, Chen, Li, Zhou, Li, Song, and
  Qi]{liu2019geniepath}
Z.~Liu, C.~Chen, L.~Li, J.~Zhou, X.~Li, L.~Song, and Y.~Qi.
\newblock Geniepath: Graph neural networks with adaptive receptive paths.
\newblock In \emph{The Thirty-Third {AAAI} Conference on Artificial
  Intelligence, {AAAI}}, pages 4424--4431, 2019.

\bibitem[Loshchilov and Hutter(2019)]{loshchilov2019decoupled}
I.~Loshchilov and F.~Hutter.
\newblock Decoupled weight decay regularization.
\newblock In \emph{7th International Conference on Learning Representations,
  {ICLR}}, 2019.

\bibitem[Loukas(2020)]{loukas2020what}
A.~Loukas.
\newblock What graph neural networks cannot learn: depth vs width.
\newblock In \emph{Proceedings of the 8th International Conference on Learning
  Representations, {ICLR}}, 2020.

\bibitem[Maron et~al.(2019{\natexlab{a}})Maron, Ben{-}Hamu, Serviansky, and
  Lipman]{maron2019provably}
H.~Maron, H.~Ben{-}Hamu, H.~Serviansky, and Y.~Lipman.
\newblock Provably powerful graph networks.
\newblock In \emph{Advances in Neural Information Processing Systems 32: Annual
  Conference on Neural Information Processing Systems 2019, {NeurIPS}}, pages
  2153--2164, 2019{\natexlab{a}}.

\bibitem[Maron et~al.(2019{\natexlab{b}})Maron, Ben{-}Hamu, Shamir, and
  Lipman]{maron2019invariant}
H.~Maron, H.~Ben{-}Hamu, N.~Shamir, and Y.~Lipman.
\newblock Invariant and equivariant graph networks.
\newblock In \emph{Proceedings of the 7th International Conference on Learning
  Representations, {ICLR}}, 2019{\natexlab{b}}.

\bibitem[Micheli and Tortorella(2023)]{micheli2023addressing}
A.~Micheli and D.~Tortorella.
\newblock Addressing heterophily in node classification with graph echo state
  networks.
\newblock \emph{Neurocomputing}, 550:\penalty0 126506, 2023.

\bibitem[Micikevicius et~al.(2018)Micikevicius, Narang, Alben, Diamos, Elsen,
  Garc{\'{\i}}a, Ginsburg, Houston, Kuchaiev, Venkatesh, and
  Wu]{micikevicius2018mixed}
P.~Micikevicius, S.~Narang, J.~Alben, G.~F. Diamos, E.~Elsen, D.~Garc{\'{\i}}a,
  B.~Ginsburg, M.~Houston, O.~Kuchaiev, G.~Venkatesh, and H.~Wu.
\newblock Mixed precision training.
\newblock In \emph{6th International Conference on Learning Representations,
  {ICLR}}, 2018.

\bibitem[Morris et~al.(2019)Morris, Ritzert, Fey, Hamilton, Lenssen, Rattan,
  and Grohe]{morris2019weisfeiler}
C.~Morris, M.~Ritzert, M.~Fey, W.~L. Hamilton, J.~E. Lenssen, G.~Rattan, and
  M.~Grohe.
\newblock Weisfeiler and leman go neural: Higher-order graph neural networks.
\newblock In \emph{Proceedings of the 33rd {AAAI} Conference on Artificial
  Intelligence, {AAAI}}, pages 4602--4609, 2019.

\bibitem[Murphy et~al.(2019)Murphy, Srinivasan, Rao, and
  Ribeiro]{murphy2019relational}
R.~L. Murphy, B.~Srinivasan, V.~A. Rao, and B.~Ribeiro.
\newblock Relational pooling for graph representations.
\newblock In \emph{Proceedings of the 36th International Conference on Machine
  Learning, {ICML}}, pages 4663--4673, 2019.

\bibitem[Oono and Suzuki(2020)]{oono2020graph}
K.~Oono and T.~Suzuki.
\newblock Graph neural networks exponentially lose expressive power for node
  classification.
\newblock In \emph{Proceedings of the 8th International Conference on Learning
  Representations, {ICLR}}, 2020.

\bibitem[Park et~al.(2019)Park, Kan, Dong, Zhao, and
  Faloutsos]{park2019estimating}
N.~Park, A.~Kan, X.~L. Dong, T.~Zhao, and C.~Faloutsos.
\newblock Estimating node importance in knowledge graphs using graph neural
  networks.
\newblock In \emph{Proceedings of the 25th {ACM} {SIGKDD} International
  Conference on Knowledge Discovery {\&} Data Mining, {KDD}}, pages 596--606,
  2019.

\bibitem[Rong et~al.(2020)Rong, Huang, Xu, and Huang]{rong2020dropedge}
Y.~Rong, W.~Huang, T.~Xu, and J.~Huang.
\newblock Dropedge: Towards deep graph convolutional networks on node
  classification.
\newblock In \emph{8th International Conference on Learning Representations,
  {ICLR}}, 2020.

\bibitem[Sato(2020)]{sato2020survey}
R.~Sato.
\newblock A survey on the expressive power of graph neural networks.
\newblock \emph{arXiv}, abs/2003.04078, 2020.
\newblock URL \url{http://arxiv.org/abs/2003.04078}.

\bibitem[Sato(2023)]{sato2023graph}
R.~Sato.
\newblock Graph neural networks can recover the hidden features solely from the
  graph structure.
\newblock In \emph{International Conference on Machine Learning, {ICML}}, pages
  30062--30079, 2023.

\bibitem[Sato et~al.(2019)Sato, Yamada, and Kashima]{sato2019approximation}
R.~Sato, M.~Yamada, and H.~Kashima.
\newblock Approximation ratios of graph neural networks for combinatorial
  problems.
\newblock In \emph{Advances in Neural Information Processing Systems 32: Annual
  Conference on Neural Information Processing Systems 2019, {NeurIPS}}, pages
  4083--4092, 2019.

\bibitem[Sato et~al.(2021)Sato, Yamada, and Kashima]{sato2021random}
R.~Sato, M.~Yamada, and H.~Kashima.
\newblock Random features strengthen graph neural networks.
\newblock In \emph{Proceedings of the 2021 {SIAM} International Conference on
  Data Mining, {SDM}}, pages 333--341, 2021.

\bibitem[Sato et~al.(2022)Sato, Yamada, and Kashima]{sato2022constant}
R.~Sato, M.~Yamada, and H.~Kashima.
\newblock Constant time graph neural networks.
\newblock \emph{{ACM} Trans. Knowl. Discov. Data}, 16\penalty0 (5):\penalty0
  92:1--92:31, 2022.
\newblock \doi{10.1145/3502733}.
\newblock URL \url{https://doi.org/10.1145/3502733}.

\bibitem[Scarselli et~al.(2009)Scarselli, Gori, Tsoi, Hagenbuchner, and
  Monfardini]{scarselli2009graph}
F.~Scarselli, M.~Gori, A.~C. Tsoi, M.~Hagenbuchner, and G.~Monfardini.
\newblock The graph neural network model.
\newblock \emph{{IEEE} Trans. Neural Networks}, 20\penalty0 (1):\penalty0
  61--80, 2009.
\newblock \doi{10.1109/TNN.2008.2005605}.
\newblock URL \url{https://doi.org/10.1109/TNN.2008.2005605}.

\bibitem[Shchur et~al.(2018)Shchur, Mumme, Bojchevski, and
  G{\"{u}}nnemann]{schur2018pitfalls}
O.~Shchur, M.~Mumme, A.~Bojchevski, and S.~G{\"{u}}nnemann.
\newblock Pitfalls of graph neural network evaluation.
\newblock \emph{arXiv}, 2018.

\bibitem[Sun et~al.(2020)Sun, Wang, Chen, Ni, Agrawal, Cui, Venkataramani,
  Maghraoui, Srinivasan, and Gopalakrishnan]{sun2020ultra}
X.~Sun, N.~Wang, C.~Chen, J.~Ni, A.~Agrawal, X.~Cui, S.~Venkataramani, K.~E.
  Maghraoui, V.~Srinivasan, and K.~Gopalakrishnan.
\newblock Ultra-low precision 4-bit training of deep neural networks.
\newblock In \emph{Advances in Neural Information Processing Systems 33: Annual
  Conference on Neural Information Processing Systems 2020, NeurIPS}, 2020.

\bibitem[Tian et~al.(2024)Tian, Song, Wang, Wang, Hu, Wang, Chawla, and
  Xu]{tian2024graph}
Y.~Tian, H.~Song, Z.~Wang, H.~Wang, Z.~Hu, F.~Wang, N.~V. Chawla, and P.~Xu.
\newblock Graph neural prompting with large language models.
\newblock In \emph{Thirty-Eighth {AAAI} Conference on Artificial Intelligence,
  {AAAI}}, pages 19080--19088, 2024.

\bibitem[Velickovic et~al.(2018)Velickovic, Cucurull, Casanova, Romero,
  Li{\`{o}}, and Bengio]{velickovic2018graph}
P.~Velickovic, G.~Cucurull, A.~Casanova, A.~Romero, P.~Li{\`{o}}, and
  Y.~Bengio.
\newblock Graph attention networks.
\newblock In \emph{Proceedings of the 6th International Conference on Learning
  Representations, {ICLR}}, 2018.

\bibitem[Wang et~al.(2019{\natexlab{a}})Wang, Qi, Lin, Cui, Jia, Wang, Fang,
  Yu, Zhou, and Yang]{wang2019semi}
D.~Wang, Y.~Qi, J.~Lin, P.~Cui, Q.~Jia, Z.~Wang, Y.~Fang, Q.~Yu, J.~Zhou, and
  S.~Yang.
\newblock A semi-supervised graph attentive network for financial fraud
  detection.
\newblock In \emph{2019 {IEEE} International Conference on Data Mining,
  {ICDM}}, pages 598--607, 2019{\natexlab{a}}.

\bibitem[Wang et~al.(2019{\natexlab{b}})Wang, Zhao, Xie, Li, and
  Guo]{wang2019knowledge}
H.~Wang, M.~Zhao, X.~Xie, W.~Li, and M.~Guo.
\newblock Knowledge graph convolutional networks for recommender systems.
\newblock In \emph{The World Wide Web Conference, {WWW}}, pages 3307--3313,
  2019{\natexlab{b}}.

\bibitem[Wang et~al.(2018)Wang, Choi, Brand, Chen, and
  Gopalakrishnan]{wang2018training}
N.~Wang, J.~Choi, D.~Brand, C.~Chen, and K.~Gopalakrishnan.
\newblock Training deep neural networks with 8-bit floating point numbers.
\newblock In \emph{Advances in Neural Information Processing Systems 31: Annual
  Conference on Neural Information Processing Systems 2018, NeurIPS}, pages
  7686--7695, 2018.

\bibitem[Wang et~al.(2019{\natexlab{c}})Wang, He, Cao, Liu, and
  Chua]{wans2019kgat}
X.~Wang, X.~He, Y.~Cao, M.~Liu, and T.~Chua.
\newblock {KGAT:} knowledge graph attention network for recommendation.
\newblock In \emph{Proceedings of the 25th {ACM} {SIGKDD} International
  Conference on Knowledge Discovery {\&} Data Mining, {KDD}}, pages 950--958,
  2019{\natexlab{c}}.

\bibitem[Wang et~al.(2021)Wang, Jin, Zhang, Yu, Zhang, and Wipf]{wang2021bag}
Y.~Wang, J.~Jin, W.~Zhang, Y.~Yu, Z.~Zhang, and D.~Wipf.
\newblock Bag of tricks for node classification with graph neural networks.
\newblock \emph{arXiv preprint arXiv:2103.13355}, 2021.

\bibitem[Wang et~al.(2022)Wang, Jin, Zhang, Yang, Chen, Gan, Yu, Zhang, Huang,
  and Wipf]{wang2022why}
Y.~Wang, J.~Jin, W.~Zhang, Y.~Yang, J.~Chen, Q.~Gan, Y.~Yu, Z.~Zhang, Z.~Huang,
  and D.~Wipf.
\newblock Why propagate alone? parallel use of labels and features on graphs.
\newblock In \emph{The Tenth International Conference on Learning
  Representations, {ICLR}}, 2022.

\bibitem[Wu et~al.(2020)Wu, Judd, Zhang, Isaev, and
  Micikevicius]{wu2020integer}
H.~Wu, P.~Judd, X.~Zhang, M.~Isaev, and P.~Micikevicius.
\newblock Integer quantization for deep learning inference: Principles and
  empirical evaluation.
\newblock \emph{arXiv}, abs/2004.09602, 2020.
\newblock URL \url{https://arxiv.org/abs/2004.09602}.

\bibitem[Xu et~al.(2019)Xu, Hu, Leskovec, and Jegelka]{xu2019how}
K.~Xu, W.~Hu, J.~Leskovec, and S.~Jegelka.
\newblock How powerful are graph neural networks?
\newblock In \emph{Proceedings of the 7th International Conference on Learning
  Representations, {ICLR}}, 2019.

\bibitem[Yang et~al.(2016)Yang, Cohen, and Salakhutdinov]{yang2016revisiting}
Z.~Yang, W.~W. Cohen, and R.~Salakhutdinov.
\newblock Revisiting semi-supervised learning with graph embeddings.
\newblock In \emph{Proceedings of the 33rd International Conference on Machine
  Learning, {ICML}}, pages 40--48, 2016.

\bibitem[Ying et~al.(2018)Ying, He, Chen, Eksombatchai, Hamilton, and
  Leskovec]{ying2018graph}
R.~Ying, R.~He, K.~Chen, P.~Eksombatchai, W.~L. Hamilton, and J.~Leskovec.
\newblock Graph convolutional neural networks for web-scale recommender
  systems.
\newblock In \emph{Proceedings of the 24th {ACM} {SIGKDD} International
  Conference on Knowledge Discovery {\&} Data Mining, {KDD}}, pages 974--983,
  2018.

\bibitem[Zeng et~al.(2020)Zeng, Zhou, Srivastava, Kannan, and
  Prasanna]{zeng2020graph}
H.~Zeng, H.~Zhou, A.~Srivastava, R.~Kannan, and V.~K. Prasanna.
\newblock Graphsaint: Graph sampling based inductive learning method.
\newblock In \emph{8th International Conference on Learning Representations,
  {ICLR}}, 2020.

\bibitem[Zhang et~al.(2023)Zhang, Luo, Wang, and He]{zhang2023rethinking}
B.~Zhang, S.~Luo, L.~Wang, and D.~He.
\newblock Rethinking the expressive power of gnns via graph biconnectivity.
\newblock In \emph{The Eleventh International Conference on Learning
  Representations, {ICLR}}, 2023.

\bibitem[Zhu and Ghahramani(2002)]{zhu2002learning}
X.~Zhu and Z.~Ghahramani.
\newblock Learning from labeled and unlabeled data with label propagation.
\newblock 2002.

\bibitem[Zou et~al.(2019)Zou, Hu, Wang, Jiang, Sun, and Gu]{zou2019layer}
D.~Zou, Z.~Hu, Y.~Wang, S.~Jiang, Y.~Sun, and Q.~Gu.
\newblock Layer-dependent importance sampling for training deep and large graph
  convolutional networks.
\newblock In \emph{Advances in Neural Information Processing Systems 32: Annual
  Conference on Neural Information Processing Systems 2019, {NeurIPS}}, pages
  11247--11256, 2019.

\end{thebibliography}
\bibliographystyle{abbrvnat}

\end{document}